%% file: main.tex
\documentclass[letterpaper, 10 pt, conference]{ieeeconf}
\IEEEoverridecommandlockouts
\usepackage{cite}
\usepackage{amsmath,amssymb,amsfonts}
\usepackage{algorithmic}
\usepackage{graphicx}
\usepackage{textcomp}
\usepackage{xcolor}
\usepackage{url}
\usepackage{placeins}
\usepackage{float}
\usepackage{tabularx,colortbl}
\usepackage{ifthen}
\usepackage{caption,subcaption}
\usepackage{amsmath}
\usepackage{booktabs}
\renewcommand{\thetable}{\Roman{table}}
\usepackage{cleveref}
\let\labelindent\relax
\usepackage{enumitem}
\usepackage{stfloats}
\usepackage{wrapfig}
\usepackage{verbatim}
\usepackage{flushend}
\def\BibTeX{{\rm B\kern-.05em{\sc i\kern-.025em b}\kern-.08em
    T\kern-.1667em\lower.7ex\hbox{E}\kern-.125emX}}

\begin{document}
\newcolumntype{P}[1]{>{\centering\arraybackslash}p{#1}}
\newcolumntype{M}[1]{>{\centering\arraybackslash}m{#1}}
\setlength{\textfloatsep}{10pt plus 1.0pt minus 3.0pt}
\setlength{\dbltextfloatsep}{10pt plus 1.0pt minus 3.0pt}
\setlength{\floatsep}{10pt plus 1.0pt minus 3.0pt}
\setlength{\dblfloatsep}{10pt plus 1.0pt minus 3.0pt}
\setlength{\intextsep}{10pt plus 1.0pt minus 3.0pt}
\input{Tex_Files/write.tex}

\bibliographystyle{IEEEtran}
\bibliography{main.bbl}

\end{document}

%% file: Tex_Files/write.tex
\title{\LARGE \bf Mutual Information-calibrated Conformal Feature Fusion for Uncertainty-Aware Multimodal 3D Object Detection at the Edge}

\author{Alex C. Stutts, Danilo Erricolo, Sathya Ravi, Theja Tulabandhula, and Amit Ranjan Trivedi
\thanks{\textbf{Acknowledgement:} This work was supported in part by COGNISENSE, one of seven centers in JUMP 2.0, a Semiconductor Research Corporation (SRC) program sponsored by DARPA. Authors are with the University of Illinois Chicago (UIC), Chicago, IL, Email: {\tt\small astutt2@uic.edu}}
}
\maketitle

\begin{abstract} 
In the expanding landscape of AI-enabled robotics, robust quantification of predictive uncertainties is of great importance. Three-dimensional (3D) object detection, a critical robotics operation, has seen significant advancements; however, the majority of current works focus only on accuracy and ignore uncertainty quantification. Addressing this gap, our novel study integrates the principles of conformal inference (CI) with information theoretic measures to perform lightweight, Monte Carlo-free uncertainty estimation within a multimodal framework. Through a multivariate Gaussian product of the latent variables in a Variational Autoencoder (VAE), features from RGB camera and LiDAR sensor data are fused to improve the prediction accuracy. Normalized mutual information (NMI) is leveraged as a modulator for calibrating uncertainty bounds derived from CI based on a weighted loss function. Our simulation results show an inverse correlation between inherent predictive uncertainty and NMI throughout the model's training. The framework demonstrates comparable or better performance in KITTI 3D object detection benchmarks to similar methods that are not uncertainty-aware, making it suitable for real-time edge robotics. 
\end{abstract}

\section{Introduction}
The rapid development of artificial intelligence (AI) capabilities, as demonstrated with image recognition and large language models (LLMs), has enabled its adoption across various domains. However, concerns about its reliability persist for safety-critical applications, including robotics. Given that the accuracy of data-driven models cannot be assured, it becomes essential not only to question \textit{what if the model is wrong?}, but also to determine \textit{how wrong} it might be by assessing its predictive uncertainties. Quantifying uncertainty in deep learning has, therefore, gained traction. Notably, data-driven models can suffer from two main types of uncertainties: \textit{epistemic} and \textit{aleatoric} \cite{kendall2017}. Epistemic uncertainty arises from inherent data variance and can often be mitigated with additional training data. Conversely, aleatoric uncertainty stems from random data distortions, such as blurriness, occlusions, and overexposure in images, and cannot be resolved merely by augmenting the training data. 

However, much of the current work has neglected considering platforms with time, cost, area, computing, and power constraints. Consequently, those existing uncertainty estimation methods, often reliant on distribution-based approximations, struggle under edge deployment due to their need for iterative sampling. Therefore, uncovering true, statistically confident uncertainties in point (mean) predictions that are intuitive and visualizable under considerable resource constraints remains challenging for critical edge robotics.

To tackle these challenges, we explore conformal inference (CI) \cite{vovk2005,shafer2007,angelopoulos2021}. Rooted in information theory and probabilistic prediction, CI has emerged as a prominent uncertainty quantification method that is simple, generalizable, and scalable \cite{manokhin_valery_2022_6467205}. Unlike conventional statistical inference, which depends on intimate knowledge of the data distribution for uncertainty estimation and is vulnerable to modeling inaccuracies, CI produces reliable, uncertainty-aware prediction intervals without distributional assumptions given a finite set of training data. CI assesses the conformity of each incoming data point to the existing dataset and formulates uncertainty intervals based on a preset coverage rate. Importantly, CI is compatible with any core model with an inherent uncertainty notion, yielding both model-agnostic and statistically sound estimations.

\begin{figure}[!t]
  \centering
  \includegraphics[width=\linewidth]{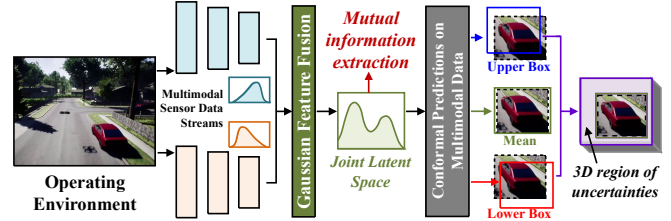}
  \caption{\textbf{Uncertainty-Aware Multimodal Inference at the Edge:} In this work, we present a generalizable, multimodal conformal inference framework for lightweight uncertainty awareness and apply it to 3D object detection. The proposed methodology is deeply rooted in information and statistical theory, allowing the framework to take full advantage of the benefits of conformal prediction in quantifying uncertainty while under considerable resource constraints.}
  \vspace{-6pt}
  \label{fig:kitti}
\end{figure}

Despite these advantages, a key limitation of CI is its tendency to provide overly cautious uncertainty estimates that may prevent a prediction model from making meaningful decisions. For example, overly conservative uncertainty estimates in autonomous navigation can lead to suboptimal path planning, such as taking longer routes than necessary. While multimodal sensors have become prevalent in various robotics tasks to enhance the robot's perception and decision-making capabilities, they present a unique opportunity for CI to optimally calibrate the predicted uncertainty estimates by exploiting mutual information (MI) of multimodal sensor data streams. MI is an information-theoretic metric that measures the dependence between the marginal distributions of two random variables through their joint distribution. In this case, it can measure how much one sensor modality explains the output and prediction from another while operating in the same environment. Thus, leveraging MI to calibrate and tighten CI's predictive uncertainty bounds while maintaining the guaranteed coverage rate is attractive.

Towards this goal, we consider 3D object detection a driving application and present a systematic framework for including MI in optimizing CI-based uncertainty bounds. 3D object detection is essential for many autonomous systems to provide a semantic understanding of their environment through identifying, localizing, and categorizing various objects. However, various propositions in our work are also generalizable to other autonomy tasks. 

Our work makes the following key contributions:
\begin{itemize}[itemsep=2pt,topsep=2pt,leftmargin=10pt]
\item We introduce a 3D object detection framework that integrates uncertainty-aware projections obtained through conformal prediction. Evaluated on the demanding 3D KITTI vision benchmark suite \cite{Geiger2012CVPR}, this framework surpasses state-of-the-art models in inference runtime while achieving a competitive accuracy. Given these attributes, our approach is particularly suited for edge robotics platforms with limited time and computational resources. 
\item We introduce a multitask loss function that can train a model to simultaneously provide point predictions and adaptive uncertainty confidence bounds that each take the form of 3D bounding boxes. The uncertainty boxes are demonstrated to enhance average precision and are combined to be more visually intuitive. Furthermore, we weight the loss function with an uncertainty-based distance metric, averaged over every dimension of each output, to influence the model to prioritize training samples that introduce more uncertainty. 

\item Integrating conformal inference with information-theoretic measures, specifically MI, we discuss a method to fuse data from multimodal sensors using a multivariate Gaussian product of latent variables in a variational autoencoder (VAE). The proposed VAE-based multimodal data fusion captures salient features of each modality and enables us to compute normalized mutual information (NMI). This, in turn, allows us to optimally calibrate the uncertainty bounds in a sample-adaptive manner. 

\end{itemize}

In Sec. II, we discuss the current art of 3D object detection. In Sec. III, we present the proposed framework of uncertainty-aware multimodal 3D object detection. Sec. IV presents the simulation results and Sec. V concludes. 

\section{Current Art on 3D Object Detection}
In this study, we focus on 3D object detection as a case study to demonstrate the efficacy of MI-based conformal feature fusion in achieving uncertainty awareness in multimodal sensing, particularly at the edge. 3D object detection is fundamental for existing and emerging robotic platforms, such as robotaxis, to understand environments comprehensively by detecting, localizing, and classifying objects. While 2D object detection offers basic object localization and recognition, 3D detection further enriches applications by adding depth and distance insights. This necessitates a sophisticated perception system, integrating diverse sensors like RGB cameras, LiDAR, and mmWave radar, which mutually enhance their performance.

For deep learning-based 3D object detection, we specifically focus on RGB camera images and LiDAR point cloud data. Prior works have developed state-of-the-art 3D object detection framework through early \cite{fpointnet2018,pointpainting2020}, intermediate \cite{pointfusion2018,mvxnet2019,MV3D2017,AVOD2018}, and late \cite{CLOCS2020} information fusion of LiDAR and camera streams, as LiDAR features are rightfully superior to camera features in assessing depth for 3D tasks \cite{Mao2023}. Early fusion improves data preprocessing and detection results, but often requires an additional network for initial image data processing, which increases inference runtime. Intermediate fusion offers deeper integration of multimodal features, which enhances bounding box prediction accuracy, but properly doing so remains an open problem due to the considerable distinctions in feature information and view points. Late fusion is more computationally efficient, but its performance is limited due to the lack of capturing the deep covariance between the modalities.

Notably, the above frameworks vary in their processing of LiDAR as well, with three primary methods identified as point-based \cite{pointnet2016,edgeconv2019,Shi2018PointRCNN3O}, grid-based \cite{voxelnet2018,pointpillars2019,SECOND2018}, and range-based \cite{lasernet2019,rapoport2021} methods. Point-based methods involve direct predictions based on downsampled points and extracted features, which has influenced many subsequent state-of-the-art works but makes it difficult to balance appropriate sampling with efficiency. Grid-based methods rasterize point cloud data into grid representations such as voxels (volumetric pixels), pillars (vertically extended voxels), or bird's-eye view (BEV) 2D feature maps, which can provide richer and more organized 3D information, potentially leading to more accurate predictions, but require more time and memory to process. Range-based methods consist of processing 2D range views (spherical projections of point clouds), which inherently contain 3D distance as opposed to simple RGB and can therefore be easily integrated with existing efficient 2D backbones but nonetheless suffer from common 2D issues (e.g., occlusion and scale variation) that exacerbate aleatoric uncertainty. Among these prior works, PointPillars \cite{pointpillars2019}, introduced in 2018, remains the fastest inference model on the 3D KITTI vision benchmark suite \cite{Geiger2012CVPR} with a 16~ms runtime. PointPillars is a LiDAR-only model that also demonstrated comparable accuracy to other state-of-the-art models published around the same time.

Most previous 3D object detection frameworks focus primarily on accuracy; there are relatively few works that have explored uncertainty quantification \cite{meyer2020,monorun2021,oleksiienko2023uncertaintyaware,zhong2020uncertaintyaware,liu2022autoregressive,feng2019}. While these works underscore the significance of uncertainty and its potential to improve performance, their methodologies largely hinge on Bayes' theorem, maximum likelihood, or coarse statistics such as standard deviation. Such methods, deeply tied to data, model, and specific assumptions (e.g., gaussianity), can face numerical instability and might not be optimal for resource-limited systems, such as edge robotics. Addressing this critical need, in this paper, we discuss an uncertainty-aware 3D object detection framework comparable to PointPillars in speed and accuracy while providing statistically rigorous and generalizable uncertainty estimations via conformal inference.

\section{Uncertainty-Aware Multimodal 3D Object Detection by Conformal Inference}
\begin{figure*}[!t]
  \centering
  \includegraphics[width=\linewidth]{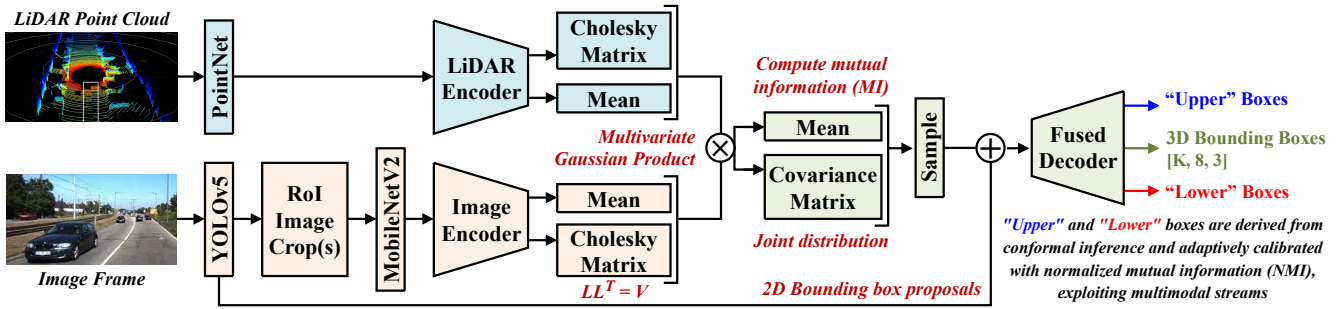}
  \caption{\textbf{Model Architecture (Section 3):} The network designed for this work utilizes a variational autoencoder (VAE) featuring dual encoders for LiDAR point cloud and RGB camera image features. To extract the multimodal features, we rely on PointNet \cite{pointnet2016}, YOLOv5s \cite{yolov5}, and MobileNetV2 \cite{howard2017}, keeping the network modular at a small expense of speed so that the feature extractors can be interchanged based on scene conditions. The information from the data streams are fused through a multivariate Gaussian product of their artificial 4D latent variables to approximate a proper covariant joint distribution. Mutual information between the multimodal features is computed using the joint mean and covariance, and a sample is extracted and concatenated with 2D bounding box proposals. Finally, a single decoder propagates the fused data to output $K$ mean 3D bounding boxes of size $[8, 3]$ along with conformal inference-based upper- and lower- bound uncertainty estimates. $K$ represents any number of detected objects.}
  \vspace{-6pt}
  \label{fig:architecture}
\end{figure*}

This section provides an overview of the proposed uncertainty-aware 3D object detection framework with RGB camera and LiDAR sensors. The model architecture is shown in Fig.~\ref{fig:architecture} and consists of a variational autoencoder (VAE) with parallel encoders for each sensor's extracted features and a single decoder that propagates information fused latent samples concatenated with 2D bounding box proposals to produce 3D bounding boxes and uncertainty bounds. To extract LiDAR features, the model relies on PointNet \cite{pointnet2016}. To obtain 2D bounding box proposals and subsequently extract camera features from cropped regions-of-interest (RoI), the model uses YOLOv5s \cite{yolov5} and MobileNetV2 \cite{howard2017}. This approach takes inspiration from PointFusion \cite{pointfusion2018}, as we opted for point-based LiDAR point cloud processing and intermediate LiDAR-camera fusion. These design choices were made primarily in consideration of efficiency and providing a solid information theoretic testbed for conformal inference.

\subsection{Feature Fusion by Multivariate Gaussian Product}
To effectively merge features extracted from RGB camera images and LiDAR point clouds, we adopt an approach inspired by \cite{ren2021}. They showed that the univariate Gaussian product of RGB and infrared image features optimally combines information from both modalities, ensuring the network remains resilient even when one data stream is suboptimal. Leveraging the Variational Autoencoder's (VAE) ability to approximate Gaussian distributions through reparameterization and Kullback-Leibler (KL) divergence of artificial latent variables (e.g., mean and variance) \cite{doersch2015}, we extend this statistical approach with a multivariate Gaussian product. Shifting from univariate to multivariate variables requires significant changes and optimizations.

Thus, our enhancement lies in operations on multivariate mean and covariance, ensuring richer representations of multimodal data. Instead of using the VAE's dual encoders for camera and LiDAR data to output variance, we utilize them to produce 4D Cholesky decompositions \cite{golub2013} of the presumed covariance matrices for each encoded feature set. A Cholesky decomposition ($L$) represents the square root of a covariance matrix and ensures symmetry and positive definiteness--two necessary criteria for subsequent matrix operations. Moreover, it encapsulates off-diagonal relationships of the latent variables, which often provide a truer representation of the covariance matrix but are commonly zeroed out in VAEs under the assumption of conditional independence.

From the Cholesky decompositions, we derive symmetric 4D covariance matrices for each set of encoded features using the matrix product \(LL^T = V\). To fuse the feature information, we utilize both latent means and covariances. We refer to \cite{Bromiley2013ProductsAC}, which explains how to compute the mean \(\mu\) and covariance \(V\) of a joint Gaussian distribution, given by equations in (1), from the product of \(n\) marginal distributions. Importantly, the equations are generalizable, suggesting that our framework can, in principle, handle an arbitrary number of sensor modalities.
\begin{subequations}
\begin{align}
\mu_{\text{joint}} &= V_{\text{joint}} \sum_{i=1}^{n} V_{i}^{-1} \mu_{i} \\
V_{\text{joint}}^{-1} &= \sum_{i=1}^{n} V_{i}^{-1}
\end{align}
\end{subequations}

Additionally, since these matrix computations involve inversion, we must address potential numerical instabilities that can ruin the approximations and cause divergence. To mitigate these concerns, we regularize the model with identity covariance and perform an eigen decomposition, denoted as \(Q\Lambda Q^{T}=V\), on the joint covariance matrix whenever a Cholesky decomposition cannot be performed. With the latter step, we can ensure positive definiteness and avoidance of near-singularity by reconstructing the matrix after we have set the non-positive eigenvalues to a small positive constant (e.g., 1$e$-6). Finally, with the proper joint mean and covariance, we can compute the mutual information (see below) between camera and LiDAR features and subsequently forward a sample from their fused distribution to the decoder along with 2D bounding box proposals.

\subsection{Uncertainty Calibration by Mutual Information (MI)}
Given the close relationship between conformal inference and information theory, we anticipate incorporating MI should improve our uncertainty-aware framework. MI quantifies the dependence between two random variables by examining the relationship between their marginal distributions and their joint distribution \cite{cover2006}. Effectively, MI assesses the uncertainty of one random variable in explaining the information of another. Previous studies have demonstrated that maximizing MI between the input feature space and latent space enhances the model's utilization of the latent representations \cite{rezaabad2020,Zhao2019InfoVAEBL}.

In this study, we utilize MI as a criterion for calibrating the conformal uncertainty intervals. To compute MI, we determine the determinants ($|\cdot|$) of the covariance matrices constructed for both the camera and LiDAR data and the covariance matrix of their combined joint distribution.
\begin{equation}
\text{MI}=\frac{1}{2}\log_{2}\left(\frac{|V_{RGB}||V_{LiDAR}|}{|V_{joint}|}\right)
\end{equation}
Afterward, we approximate the Shannon entropy \cite{cover2006} of the two feature sets' covariances with (3) and use them to normalize the MI (i.e., compute NMI \cite{vinh2010}) to be within the range of [0,1] with (4).
\begin{equation}
\text{H}=\frac{1}{2}\log_{2}\left((2\pi e)^4|V|\right)
\end{equation}
\begin{equation}
\text{NMI}=\frac{2\text{MI}}{\text{H}_{RGB}+\text{H}_{LiDAR}}
\end{equation}

It is important to note that, in theory, MI is upper bounded by the maximum of the the Shannon entropies of the random variables involved. However, because the VAE's latent random variables typically have unbounded support (because activation functions such as ReLU and others have unbounded ranges), it is possible to run into stability issues where a network could continue optimizing its parameters leading to divergent MI estimates. To fix this, we add the $softsign()$ activation function shifted by +1 to bind the latent variables to the range $[0,2]$ and stabilize the network. This activation function resembles the hyperbolic tangent but is less steep and therefore saturates slower. In the next subsection, we discuss the placement of the NMI metric into the loss function.

\subsection{Uncertainty Weighted Loss by Conformal Inference (CI)}
CI offers a model-agnostic method for uncertainty quantification that seamlessly integrates with any foundational model possessing intrinsic uncertainty measures, such as quantile regression. The intervals guarantee marginal coverage of the truth based on a user-defined coverage rate \cite{vovk2005,shafer2007,angelopoulos2021}. Marginal coverage represents the average probability, taken over all considered samples, that true values will fall within the prediction intervals. It is analytically guaranteed by using a portion of the training data as a calibration set to compute conformity scores of new observations to prior information, which are used to calibrate the uncertainty intervals. 

The conformalized joint prediction (CJP) method presented in our prior work \cite{stutts2023lightweight} demonstrated a unique form of multivariate cross-conformal inference where a model is jointly trained to output point (mean) predictions and conditional quantiles that serve as upper and lower prediction bounds, capturing true aleatoric and epistemic uncertainty. To construct the prediction bounds, the method requires calibration of the sample data during training so as to guide the model to center predictions and maintain marginal coverage. Cross-conformal inference involves performing a number of calibration steps over all of the training data, striking a balance between the statistical efficiency of full-conformal prediction and the speed of split-conformal prediction. The method performs the calibrations dynamically over the randomized training batches as part of a multi-task loss function that simultaneously prioritizes reconstruction, KL divergence, and uncertainty interval centeredness, tightness, and coverage. As a result, it is shown that the intervals are highly tunable, flexible, and adaptive.

We make the following impactful modifications to the loss function presented in our prior work. First, we weight the reconstruction loss with a small uncertainty penalty to guide the network to prioritize resolving higher uncertainty in certain training batches. Secondly, we regularize the KL divergence with 4D covariance instead of variance. Finally, we dynamically tune the balance between interval sharpness (i.e., uncertainty distance) and marginal coverage with normalized mutual information (NMI). For completeness, the loss function is provided as:
\begin{equation}
\begin{split}
\mathcal{L}_{Total} & = \text{SmoothL1}_{loss}(y,\hat{y})\times(1 + 0.01U) \\
& + \text{KL}_{div}(\mu_{joint},V_{joint}) \\
& + \text{INTSCORE}_{loss}(y,\mathit{Q_{l},Q_{h}},\{\mathit{\alpha_{l},\alpha_{h}}\}) \\
& + \text{COMCAL}_{loss}(y,p^{cov}_{avg},\mathit{Q_{l},Q_{h},NMI})
\end{split}
\end{equation}
where
\begin{equation}
\text{KL}_{div}=\frac{1}{2}\left(Tr(V)+\mu_{joint}\mu_{joint}^{T} - 4 - \text{log}(|V|)\right)
\end{equation}
\begin{equation}
\begin{split}
\text{INTSCORE}_{loss} & =(Q_{h}-Q_{l}) + \frac{2}{\alpha}(Q_{l}-y)\mathbb{I}\{y < Q_{l}\} \\
     & + \frac{2}{\alpha}(y-Q_{h})\mathbb{I}\{y > Q_{h}\} \\
\end{split}
\end{equation}
\begin{subequations}
\begin{equation}
\text{COMCAL}_{loss} = (1-NMI)\times\text{CAL}_{obj}+NMI\times\text{SHARP}_{obj}
\end{equation}
and
\begin{equation}
\begin{split}
& \text{CAL}_{obj} = \\ 
& \mathbb{I}\{p^{cov}_{avg} < p\} \times \frac{1}{N}\sum_{i=1}^{N}[(y_{i}-Q_{l,h}(x_{i}))\mathbb{I}\{y_{i} > Q_{l,h}\}]~+ \\ 
& \mathbb{I}\{p^{cov}_{avg} > p\} \times \frac{1}{N}\sum_{i=1}^{N}[(Q_{l,h}(x_{i})-y_{i})\mathbb{I}\{y_{i} < Q_{l,h}\}] 
\end{split}
\end{equation}
\begin{equation}
\begin{split}
\text{SHARP}_{obj} & = \mathbb{I}\{p \leq 0.5\} \times \frac{1}{N}\sum_{i=1}^{N}Q_{l}(x_{i}) - Q_{h}(x_{i}) \\ 
& +~\mathbb{I}\{p > 0.5\} \times \frac{1}{N}\sum_{i=1}^{N}Q_{h}(x_{i}) - Q_{l}(x_{i}) 
\end{split}
\end{equation}
\end{subequations}

In the equations, $x$ represents input samples, $y$ represents 3D bounding box labels, $\hat{y}$ represents predictions, $U$ represents a singular uncertainty distance metric calculated by averaging the prediction interval length of each output dimension per training batch, $Q_{h}$ and $Q_{l}$ are each dimension's upper and lower quantile estimates used to calculate $U$, $\alpha_{h}$ and $\alpha_{l}$ are the 95$^{th}$ and 5$^{th}$ percentile coverage bounds that assert $Q_{h}$ and $Q_{l}$, $p$ is the chosen marginal coverage rate ($\alpha_{h}-\alpha_{l}=$ 90\%), $\mathbb{I}$ is the indicator function, $Tr()$ is the trace function, and $p^{cov}_{avg}$ is the estimated probability that the label values lie within $[\mathit{Q_{l},Q_{h}}]$, averaged over the randomized training batches. $Q_{l,h}$ is meant to indicate that $\text{CAL}_{obj}$ is computed separately for both $Q_{l}$ and $Q_{h}$ and then added together.

Focusing on the two lesser-known loss components---$\text{INTSCORE}_{loss}$ is used to influence the model to maintain centered quantile intervals while $\text{COMCAL}_{loss}$ is used to control the balance between minimizing the uncertainty intervals and increasing marginal coverage, as reflected in the sub-objectives $\text{CAL}_{obj}$ and $\text{SHARP}_{obj}$. Notably, we insert NMI from Section III-B, averaged in each training batch, into $\text{COMCAL}_{loss}$ to dynamically control the calibration balance during training as opposed to setting a static value. Intuitively, the model is influenced to be less uncertain when the MI between the RGB camera and LiDAR features is high. Therefore, uncertainty and MI should be inversely correlated.

\section{Results and Discussions}
This section details our observations from applying the framework presented in Section III to 3D object detection involving RGB cameras and LiDAR point cloud inputs. Towards this, the primary goal of the proposed framework is to enable lightweight, conformalized uncertainty awareness while including principles of entropy, MI, and feature fusion based on a multivariate Gaussian product. This combination of theories is used to improve the model's uncertainty estimates through CI while operating under edge device constraints. Notably, uncertainty estimation can become difficult and unstable for a task such as 3D object detection, where there is a varied number of multivariate objects to be assessed per input sample. Therefore, taking an approach deeply rooted in information and statistical theory is imperative. 

\begin{figure}[!t]
\centering
    \includegraphics[width=\columnwidth]{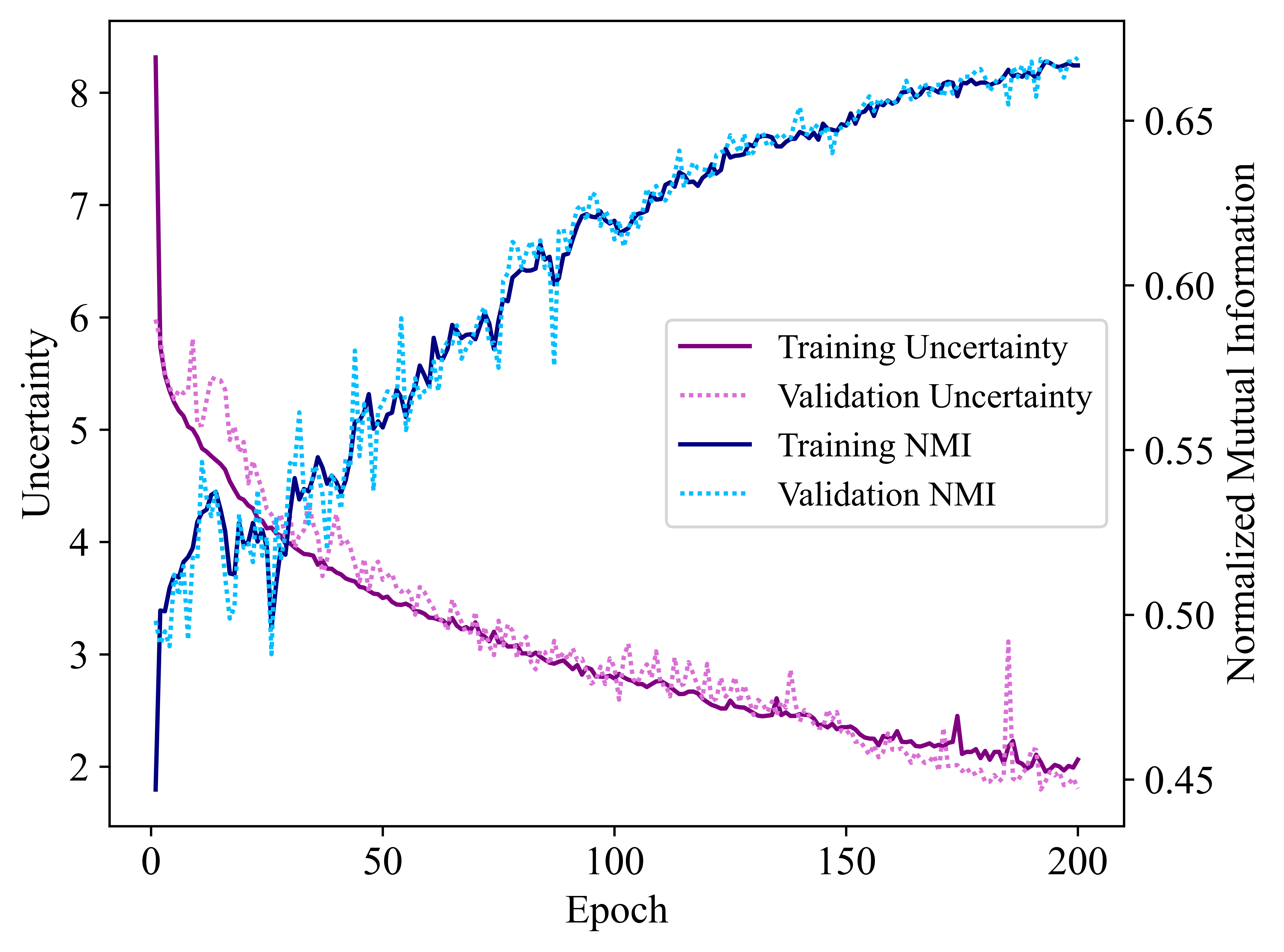}
    \caption{\textbf{Average Uncertainty and Normalized Mutual Information (NMI) vs. Epoch:} Explicit uncertainty and NMI obtained from performing conformal inference and intermediate feature fusion are averaged across training batches in each epoch. Uncertainty and NMI are inversely correlated, influencing the model to be more confident in predictions when mutual information is high and \textit{vice versa}.}
    \label{fig:uncertainties_NMI}
    \vspace{-6pt}
\end{figure}

As projected in Section III-B, it is shown in Fig.~\ref{fig:uncertainties_NMI} that the average uncertainty and normalized mutual information (NMI) obtained via conformalized feature fusion are inversely correlated over the duration of training the model described in Fig.~\ref{fig:architecture}. It is important to note that while mutual information is static given a discrete input feature space, here we are deriving it from artificial latent representations that are optimized during training. Hence, the value of NMI can change during training as the embedded information is better understood. While the estimated NMI increased between the camera and LiDAR data, the overall uncertainty in the predictions decreased. This uncertainty metric $U$ is used to weight the $\text{SmoothL1}$ reconstruction loss, while the NMI is used to calibrate this uncertainty in (8a). To the best of our knowledge, this is the first work demonstrating a stable combination of explicitly translatable uncertainty weighting and mutual information in a multitask loss function where both influence each other.

Table~I quantitatively compares our proposed framework to similar works predicting 3D bounding boxes for cars in the seminal KITTI 3D detection dataset. The easy, moderate, and hard percentage scores are of average precision in 3D bounding box regression ($AP_{3D}$), which is based on precision-recall calculations with an intersection-over-union (IoU) threshold of 0.7 in various scene conditions. Most metrics are taken directly from the KITTI source website, where the various referenced models have been submitted for result reproduction. From the table, our model is approximately 38\% faster than PointPillars without suffering an equal accuracy loss, making it suitable for edge robotics. The runtime metrics we provide are adjusted for hardware performance differences, given PointPillars used an NVIDIA GTX 1080 Ti desktop, and we used an NVIDIA RTX 4090 laptop. Unlike the other works, the model maintains a relatively consistent accuracy across each benchmark case, a unique result of using a VAE and conformal inference. 

\begin{table*}[t]
\setlength{\tabcolsep}{1pt}
\renewcommand{\arraystretch}{1.3}
\centering
\small
\begin{tabular}{M{24mm} M{20mm} M{19mm} M{19mm} M{19mm} M{12mm} M{19mm} M{19mm} M{19mm}}

\multicolumn{9}{c}{\textbf{Table I: Comparison of Proposed 3D Object Detection Framework to Similar Work on KITTI Cars $(AP_{3D})$}} \vspace{5pt}\\
\hline
\textbf{Reference} & \textbf{Modality} & \textbf{LiDAR Rep.} & \textbf{Fusion Type}  & \textbf{Uncertainty} & \textbf{Runtime (ms)} & \textbf{Easy (\%)}  & \textbf{Mod. (\%)}  & \textbf{Hard (\%)} \\
\hline
PointFusion \cite{pointfusion2018} & Cam+LiDAR & Points & Intermediate & No & -- & 74.71 & 61.24 & 50.55 \\
ContFuse \cite{contfuse2018} & Cam+LiDAR & Grid (BEV) & Intermediate & No & 60 & 83.68 & 68.78 & 61.67 \\
MVX-Net \cite{mvxnet2019} & Cam+LiDAR & Grid (Voxels) & Intermediate & No & -- & 83.2 & 72.7 & 65.2 \\
EPNet \cite{huang2020epnet} & Cam+LiDAR & Points & Intermediate & No & 100 & 89.81 & 79.28 & 76.40 \\
MMF \cite{MMF2019} & Cam+LiDAR & Grid (BEV) & Intermediate & No & 100 & 89.05 & 82.50 & 77.59 \\
MV3D \cite{MV3D2017} & Cam+LiDAR & Multiple & Intermediate & No & 360 & 74.97 & 63.63 & 54.00 \\
3D-CVF \cite{3DCVF2020} & Cam+LiDAR & Grid (BEV) & Intermediate & No & 60 & 89.20 & 80.05 & 73.11 \\
AVOD \cite{AVOD2018} & Cam+LiDAR & Grid (BEV) & Intermediate & No & 100 & 83.07 & 71.76 & 65.73 \\
CLOCs \cite{CLOCS2020} & Cam+LiDAR & Multiple & Late & No & 100 & 89.16 & 82.28 & 77.23 \\
PointPillars \cite{pointpillars2019} & LiDAR & Grid (Pillars) & Intermediate & No & 16 & 82.58 & 74.31 & 68.99 \\
\hline
\textbf{Ours} & Cam+LiDAR & Points & Intermediate & Yes & \textbf{9.87}$^*$ & 62.84 & 58.66 & 60.89 \\
\textbf{Ours w/ NMI-calibrated Uncertainty} &  &  &  &  & 14.82$^*$ & 87.64 (MAU=3.52) & \textbf{89.83} (MAU=3.59) & \textbf{92.26} (MAU=3.62) \\
\hline
\multicolumn{9}{l}{$^*$These models were characterized on an NVIDIA RTX 4090 laptop; the metrics are adjusted for an NVIDIA GTX 1080 Ti desktop.} \\
\hline
\end{tabular}
\end{table*}

\begin{figure*}[!t]
  \centering
  \includegraphics[width=\linewidth,height=6cm]{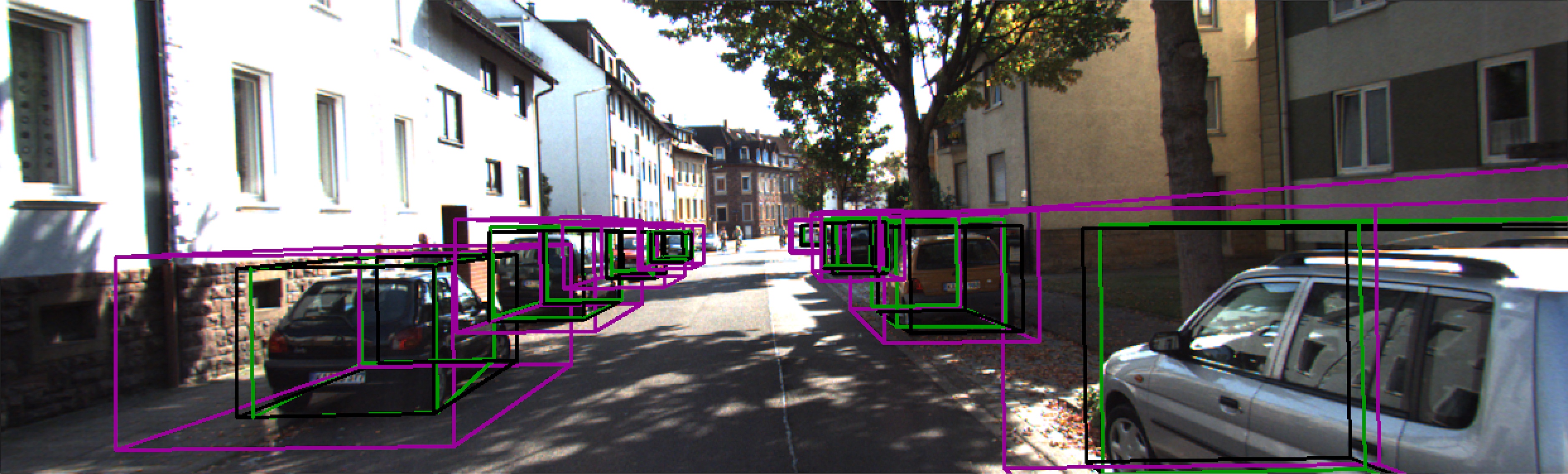}
  \caption{\textbf{Uncertainty in 3D Bounding Box Regression:} Ground truth (black), predicted (green), and uncertainty (purple) 3D bounding boxes are visualized in a sample KITTI image of 8 cars with various occlusion and truncation status. As described in Section III-C, the uncertainty boxes represent a combination of upper- and lower-bound conditional quantiles obtained via conformal inference.}
  \vspace{-6pt}
  \label{fig:kitti_sample}
\end{figure*}

Furthermore, by factoring the marginal coverage of the upper- and lower-bound uncertainty boxes calibrated with NMI into the IoU calculations, the average precision increased by at least 39\%. Accordingly, we propose an entirely new evaluation metric, \textit{mean average uncertainty} (MAU), to track the combined average uncertainty in predicting each corner of $K$ 3D bounding boxes (i.e., $\frac{1}{K}\sum_{i=1}^{K}u_{i}$). A key observation here is that, with uncertainty included, the average precision slightly increased with more difficult predictions while MAU also increased. This indicates that the model appropriately prioritized predictions where uncertainty was greater. However, it is worth noting that there are fewer annotations in the moderate and hard cases, so the model has fewer chances of being imprecise compared to the easy case. Overall, we show that robust uncertainty-awareness can improve the reliability of a model's predictions in making critical decisions and considerably improve accuracy. By maintaining a generalizable methodology, our work can be integrated to improve metrics in other models for various tasks.

Fig.~\ref{fig:kitti_sample} provides a qualitative assessment of the uncertainty in predicting the 3D bounding boxes. We display the ground truth box in black, the predicted box in green, and a combined uncertainty box in purple that encompasses the upper- and lower-bound boxes. This level of accuracy and precision in estimating and visualizing uncertainty in 3D object detection has not been demonstrated previously. A primary benefit of such assessment is that even if the model appears to be predicting well, a large uncertainty estimate can direct it to assert caution appropriately, such as when the sensors are not performing well or are impaired externally.

\section{Conclusion}
We presented a novel framework for quantifying, calibrating, and leveraging true uncertainty in multimodal inference at the edge. The proposed methodology, applied to 3D object detection, includes conformal inference, elements of information theory, and Gaussian feature fusion. Our research demonstrates that integrating uncertainty awareness not only increases reliability of data-driven deep learning, but also improves prediction accuracy and precision. The approach is both generalizable and scalable, allowing it to be adapted to any task or dataset where uncertainty awareness should be considered, especially when under considerable resource constraints such as in edge robotics. The integration of information theory and conformal inference offers benefits that extend beyond individual results in the deep learning domain.